\documentclass[conference]{IEEEtran}
\IEEEoverridecommandlockouts
\usepackage{amsmath,amssymb,amsfonts}
\usepackage{algorithmic}
\usepackage{graphicx}
\usepackage{textcomp}
\usepackage{xcolor}
\usepackage{booktabs}
\usepackage{biblatex}
\usepackage{siunitx}
\def\BibTeX{{\rm B\kern-.05em{\sc i\kern-.025em b}\kern-.08em
    T\kern-.1667em\lower.7ex\hbox{E}\kern-.125emX}}

\AtBeginEnvironment{quote}{\itshape}

\begin{document}

\title{Multi-Modal Semantic Inconsistency Detection in Social Media News Posts\\

\thanks{This research was supported by the DARPA Semantic Forensics program. Funding for compute resources was provided by Google Cloud.}
}

\author{
\IEEEauthorblockN{Scott McCrae,
Kehan Wang, 
and Avideh Zakhor}
\IEEEauthorblockA{Department of Electrical Engineering and Computer Sciences\\
University of California, Berkeley\\
% Berkeley, California, USA\\
Email: mccrae@berkeley.edu,
wang.kehan@berkeley.edu,
avz@berkeley.edu}
}

\maketitle

\begin{abstract}
As computer-generated content and deepfakes make steady improvements, semantic approaches to multimedia forensics will become more important.
In this paper, we introduce a novel classification architecture for identifying semantic inconsistencies between video appearance and text caption in social media news posts. 
We develop a multi-modal fusion framework to identify mismatches between videos and captions in social media posts by leveraging an ensemble method based on textual analysis of the caption, automatic audio transcription, semantic video analysis, object detection, named entity consistency, and facial verification. 
To train and test our approach, we curate a new video-based dataset of 4,000 real-world Facebook news posts for analysis. 
Our multi-modal approach achieves 60.5\% classification accuracy on random mismatches between caption and appearance, compared to accuracy below 50\% for uni-modal models. Further ablation studies confirm the necessity of fusion across modalities for correctly identifying semantic inconsistencies.
\end{abstract}

\begin{IEEEkeywords}
Multi-modal, disinformation, social media, forensics, fusion
\end{IEEEkeywords}

\section{Introduction}

There has been a great deal of attention on misinformation and deepfakes recently, especially with regards to the ongoing COVID-19 pandemic and the 2020 US Presidential election. There are a variety of methods for detecting both manipulated media, such as Photoshopped images, and data which is machine-generated, such as images from generative adversarial networks (GANs). 
However, these tools tend to focus on a single modality, such as imagery, and look for clues that the image has been manipulated using statistical methods or by leveraging metadata. While these tools are indisputably useful, we are interested in investigating multi-modal analysis, where we attempt to detect manipulations or misinformation using semantic clues from a variety of modalities. 

The use of multiple modalities allows us to reason about the semantic content of each source. For instance, a caption describing an out-of-control protest would be inconsistent with a video of a candle-light vigil, and a video of a reporter in the midst of a hurricane in Florida would be inconsistent with a news article on the effects of deforestation in the Amazon. On their own, neither modality is manipulated, but together they represent an inconsistency. This might model the threat of "cheapfakes," where an attacker lazily sources pairs of material to output misinformation at scale; an attacker attempting to misrepresent some original source; or an attacker with one or more undetectably altered modalities generated by a system unaware of high-level semantic consistency. While current methods are able to detect GAN-based images or deepfakes and text generated from language models, such generation techniques may continue to improve and begin to fool uni-modal detection approaches.

In this paper, we introduce a novel classification architecture for identifying semantic inconsistencies between video appearance and text caption in social media news posts.
To analyze the semantic alignment of videos and captions, we need three main ingredients. First, and most importantly, we need pristine data as ground truth. 
Second, we need to extract semantic feature representations from each modality and its constituents, such as transcripts and named entities. 
Third, we need to jointly reason about semantic content. In the following sections, each of these components will be addressed in turn. 
Section \ref{section:related_works} describes related work in natural language processing, computer vision, and multi-modal analysis. Section \ref{section:data} describes our data collection and pre-processing methods. Section \ref{section:experiments} describes experimental results and ablation studies. Section \ref{section:conclusion} provides a conclusion and a discussion of future work for this project.

\section{Related Works}
\label{section:related_works}

The field of natural language processing has seen a rapid shift in recent years towards transformer-based methods, introduced in \cite{vaswani2017attention}, with large language models achieving state of the art performance \cite{devlin2019bert, liu2019roberta, radford2019language_gpt2, brown2020language_gpt3}. 
Machine learning in computer vision has been dominated by convolutional methods, with 2D methods such as ResNet \cite{he2015resnet} becoming standard backbone networks. Several later works have extended 2D convolutional networks to process videos \cite{inception3d, hara3dcnns, xie2018rethinking}. 
Approaches such as \cite{inception3d} extend convolution into three dimensions, while \cite{xie2018rethinking} introduces separable computations over the spatial and temporal domains to increase efficiency.
\cite{miech19endtoend} adapts \cite{xie2018rethinking} to include text embeddings which are jointly learned with video embeddings, and is trained on a very large corpus of instructional videos \cite{miech19howto100m}. 
Recent research has shown promising results adapting transformer methods to process videos \cite{bertasius2021spacetime}, opening the door to processing video clips which are longer than a few seconds. 

\begin{figure*}[h]
  \centering
    \includegraphics[width=5.5in]{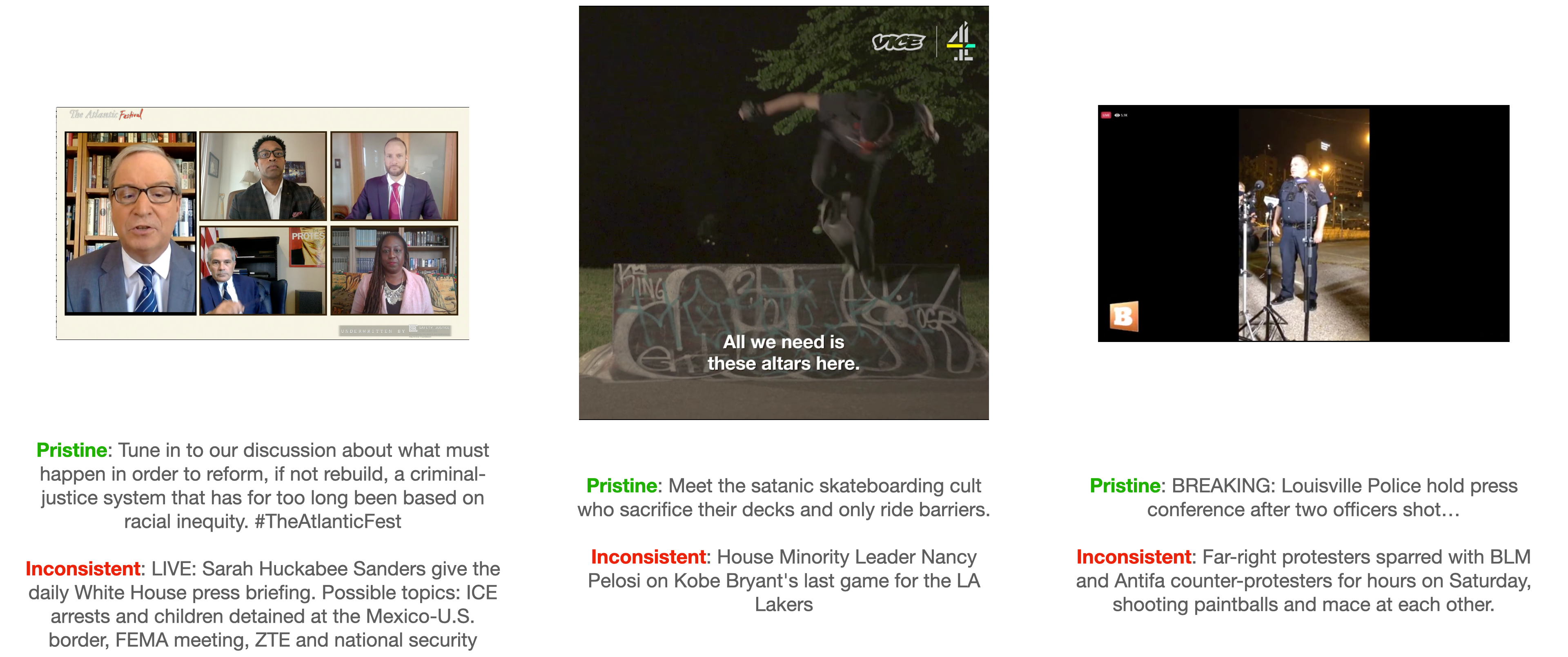}

  \caption{Example videos and captions from our dataset.}
  \label{exampledata}
\end{figure*}

Research in multi-modal learning with text and imagery has demonstrated the efficacy of learning modality-specific embeddings \cite{habibian_video2vec}. New methods have been developed with the goal of leveraging transformers to jointly process text and imagery \cite{lu2019vilbert, su2020vlbert, li2019unicodervl, tan2019lxmert}. 
\cite{Luo2020UniVL} extends joint text and image transformer-based methods to process text and video clips. \cite{li-etal-2020-hero} employs cross-modal transformers with video frame and text embeddings for multi-modal learning.

A variety of methods have been introduced recently for detecting computer-generated content and semantic inconsistencies. \cite{zellers2019grover} detects neural fake news by modeling a joint distribution over a news article's domain, date, authors, headline, and body. \cite{wang2019cnngenerated} demonstrates the relative ease of detecting GAN-generated images from a variety of state-of-the-art generators at the time of publication.
\cite{tanDIDAN2020} checks for consistency between a news article and its images and captions. 
\cite{shekhar-etal-2017-foil} attempts to identify and attribute inconsistencies between images and their captions. \cite{luo2021newsclippings} introduces and evaluates detection methods on a new dataset for the task of identifying various semantic inconsistencies between images and captions.

\section{Data and Representation}
\label{section:data}

\subsection{Dataset Design}
We construct our dataset using raw data accessed via CrowdTangle \cite{crowdtangle}, a public insights tool owned and operated by Facebook. The platform can surface public Facebook posts, including sources such as posts by celebrities and news outlets. It does not include paid advertisements unless they began as organic, non-paid posts that were subsequently “boosted” using Facebook’s advertising tools. It also does not include activity on private accounts, or posts made visible only to specific groups of followers.

We used the historical data function of the platform to construct our dataset. With the historical data feature, we downloaded all public Facebook posts which had videos in the last decade from the US General Media group, for a total of 647,009 posts. This list of organizations was curated by CrowdTangle, and ranges from large, relatively non-partisan sources such as The Associated Press to smaller, more partisan sources such as Breitbart News.

While CrowdTangle provides access to large amounts of Facebook posts, it has two limitations that impact this project. First, it does not provide labels for whether or not a post contains misinformation. Second, since it does not provide video files, they must be scraped from Facebook using other tools. 
Therefore we used CrowdTangle to source posts to scrape and used the open-source \texttt{youtube-dl} tool \cite{youtubedl} to scrape video files.
Due to this limitation, we were only able to scrape a sample of 4,651 videos.

To construct a labelled dataset for multi-modal semantic alignment, we treat the original caption-video post pairs as pristine examples and randomly swap in new captions from other posts to generate inconsistent examples. Examples are shown in Figure \ref{exampledata}.
In this manner, a pristine example features a real-world video and a real-world caption which were intended to relate to each other by the organization which posted them. We assume that pristine examples are semantically consistent across modalities. 
An inconsistent example still features a real-world video and caption, except the video and caption are not taken from the same post. In an inconsistent example, the caption is the only modality which is manipulated, i.e. swapped. For the additional modalities described in following subsections, such as a video's transcript and Facebook reactions, each example video is always paired with its matching transcript and reactions.
This assumes that a random swap of caption would result in some amount of semantic mismatch between the new caption and the original video. In practice, half of the examples in our dataset are pristine and half are inconsistent.

\begin{figure*}[h]
  \centering
    \includegraphics[width=5.75in]{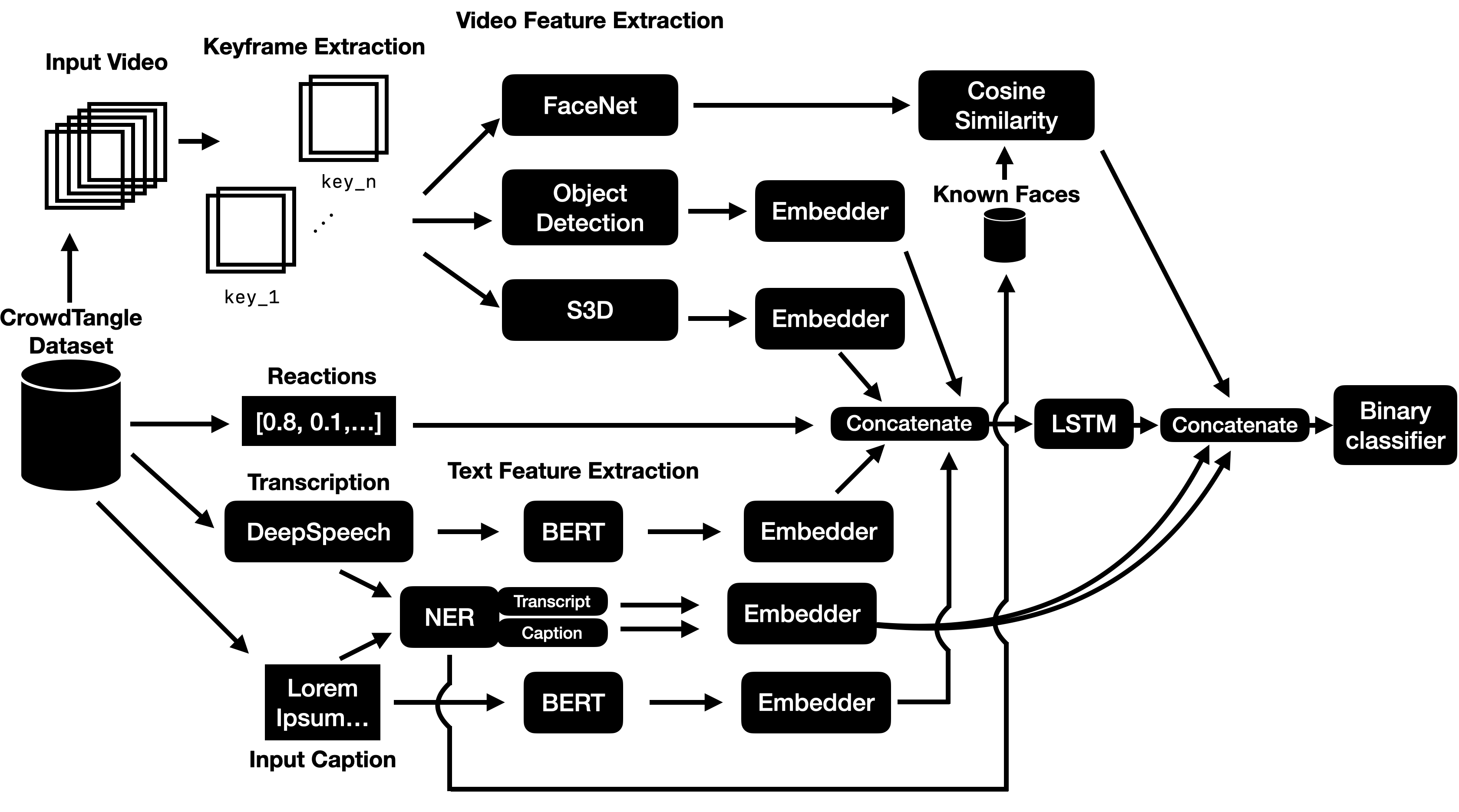}
  \caption{Our semantic inconsistency detection architecture. Modality-specific feature extraction is run in parallel, and features representing the content of each modality are concatenated with facial verification features in order to perform classification. In practice, expensive operations such as transcription are run in advance.}
  \label{pipeline}
\end{figure*}

We opt to perform swaps on real-world captions rather than creating inconsistencies by generating captions using large language models. This avoids reducing the problem of identifying semantic inconsistencies across modalities to the problem of detecting whether or not a caption is synthetically generated.
Although some real news posts may include synthetically generated text, such as short reports on financial news \cite{nytrobotnews}, we do not attempt to filter out posts which might contain synthetic text. If such synthetic posts are present, they would not be correlated with semantic inconsistency labels due to our random swapping approach.

Our chosen task of detecting caption and video appearance inconsistency is challenging because of the abstract relationships between captions, videos, and other modalities. 
The captions in our dataset do not represent dense descriptions of video clips, nor are they necessarily a literal description of a video. Our transcripts are noisy due to automatic generation, and are not guaranteed to be faithful representations of what is said. We do not have audio descriptions of videos.
Our videos cover a wide range of styles and subjects, and are not necessarily well-lit and well-produced, as one might expect in datasets with movie clips.
However, the random swapping approach does make this task easier than some more adversarial swapping strategies. 
We hope to strike a balance between perceived human difficulty and the challenge of learning abstract associations between modalities from a small set of noisy data. 

\subsection{Video Pre-Processing}
\label{subsection:video_preprocessing}

After collecting video data, we take several steps to standardize formats and to prepare the files for input to our system. Figure \ref{pipeline} illustrates how data flows through our model. Each video is transcoded to a constant resolution of 256$\times$256 pixels and a constant frame rate of 10 frames per second. All files are converted to mp4 videos, regardless of the original format. Audio is left unchanged. Video transcoding is handled using FFmpeg \cite{ffmpeg}.

Because videos are scraped at random from Facebook, there is a very wide range of video lengths, styles, and subjects. In our dataset, the minimum video length is 1 second, the maximum length is 14 hours, and the mean is 8.5 minutes. To handle the long and variable video lengths, we adopt a keyframe-based approach. Each video is broken up into a sequence of 32-frame-long clips, with each clip beginning at a keyframe.

A keyframe is intended to be a point in the video where there is a change in scene or camera angle. 
These keyframes should be well-aligned to the starts of semantically consistent clips.
In practice, we identify keyframes as timestamps in a video where the FFmpeg \cite{ffmpeg} scene detection filter is triggered, with the scene detection threshold set at $0.4$. 
If no keyframes are detected, which might be the case with very short videos or videos which are all one shot, we create placeholder keyframes every $3.2$ seconds, corresponding to 32 frames. In this manner, a clip with no detected keyframes is split into 32-frame-long clips every 32 frames. We choose to use 16 keyframes per video, taking into account that 73\% of videos in our dataset have at most 16 keyframes. We did not observe a significant difference in performance between using 8 or 16 keyframes.

Every video is transcribed using the DeepSpeech \cite{hannun2014deepspeech} transcription system. 
Before passing a video's audio stream into DeepSpeech, we transcode it using FFmpeg to the PCM signed 16-bit little-endian format with a sample rate of $16 \si{\kilo\hertz}$, apply a highpass filter with cutoff $200 \si{\hertz}$, and apply a lowpass filter with cutoff $3 \si{\kilo\hertz}$. 
Using these filters, the generated transcripts are generally faithful and fluent, although they are imperfect and tend to misspell named entities. Below is an excerpt from an example audio transcript with typos generated using DeepSpeech:

\begin{quote}
the fourth democratic presidential debate wrapped up in ohio on tuesday minnesota senator amicable no time getting back on the campaign trail she picked off with a tour of new hampshire traveling to all ten counties and just thirty hours overcasting a wave of support after snagging the spotlight on tuesday night going head to head against fortune elizabeth warehouses not even the billionaire to protect billionaire wreaking time locked in and more than thirteen minutes that prefer in sir behind warren and former vice president joined some people like what they heard on twitter cobhouse received one point one million dollars in campaign donations in the twenty four hours after the debate …
\end{quote}

While our transcripts are mostly correct, they tend to include misspelled names and other misidentified words. In this case, misspelled names include "amicable," "warehouses," and "cobhouse." The correct names are "Amy Klobuchar," "Warren," and "Klobuchar." These errors make it difficult to compare named entities in captions and transcripts, as transcript typos might not correspond to common human mistakes which might be corrected by spell-check methods.

While some videos provide closed captions, we use automatically generated transcripts uniformly across our dataset to avoid introducing any linguistic biases in the fluency or style of transcripts from different sources.

\subsection{Named Entity Verification}
\label{ssection:facialverification}
In this section we describe our approaches to verifying named entities using facial verification and text-based comparison of names in captions and audio transcripts.
\subsubsection{Facial Verification}
We implement facial verification for named entities in order to check semantic consistency between modalities. This subsection will describe the implementation of our facial verification system.

We define facial verification in our context as checking whether or not people named in the caption of a video actually appear in the video. To accomplish this, we need to identify people in captions and build a database of representations for them. 
People are identified by using the named entity recognition (NER) feature available in the spaCy \cite{spacy} natural language processing library. 
Using spaCy's \texttt{en\_core\_web\_trf} language model, which implements RoBERTa \cite{liu2019roberta}, we run NER on our dataset of captions, and take all strings with the \texttt{PERSON} label as names of people. These strings are compiled into a set of people whose names appear in our dataset.

\begin{figure}[h]
  \centering
    \includegraphics[width=3.45in]{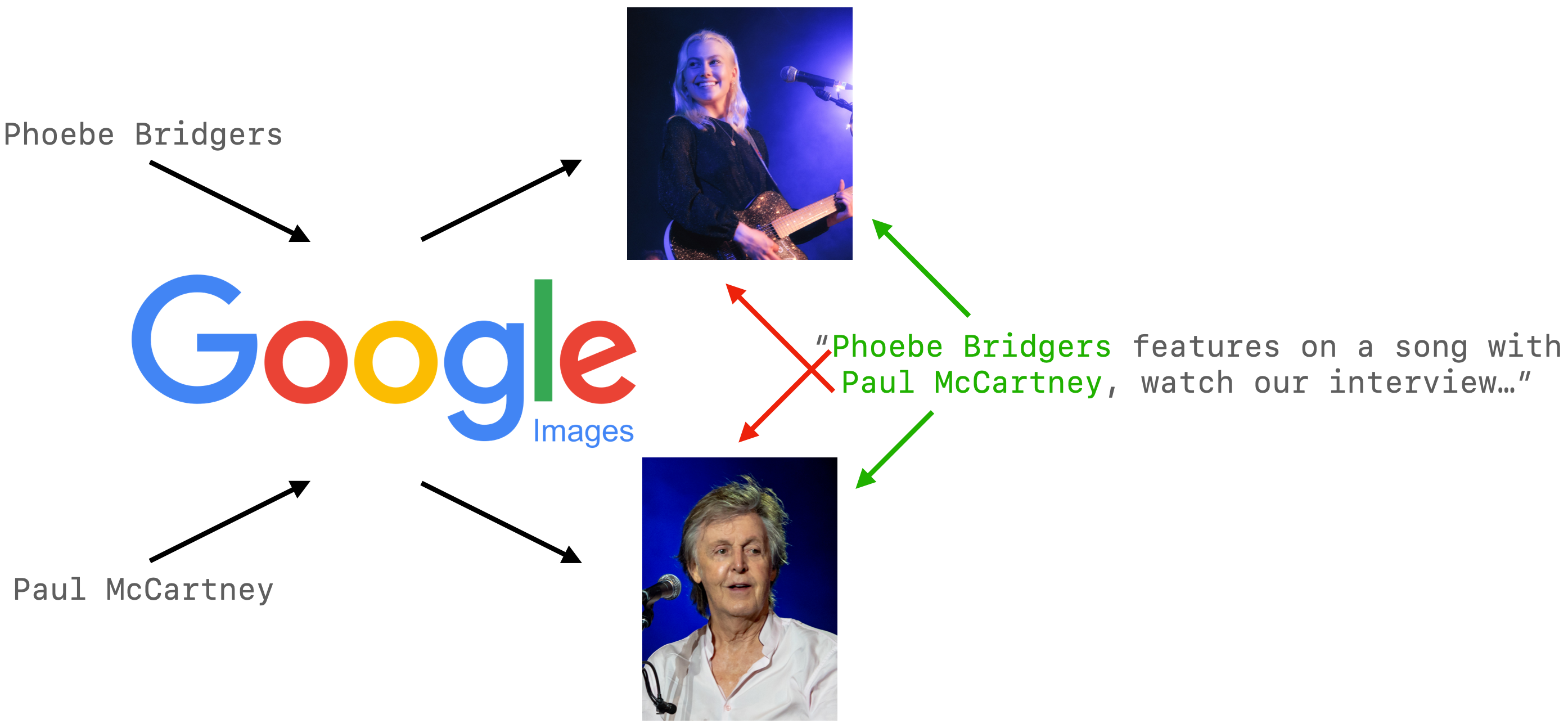}
    
  \caption{Representations of named entities in a caption, generated by querying Google Images, are compared against frames of a video.
  In this example, musicians Phoebe Bridgers and Paul McCartney could be verified by checking for their faces in the video, although there will also be apparent mismatches between each person's name and the other face present.
  Images courtesy \cite{wikipedia:bridgers, wikipedia:mccartney}. Best viewed in color.
  }
  \label{figure:google_query}
\end{figure}

Once all named people are identified, we compute a representation for each person. To this end, we query Google Images for the top 10 results for each name. 
These images are considered ground-truth references for how each named entity should appear, as shown in Figure \ref{figure:google_query}. 
Having multiple images per name allows our dataset to contain examples of each person with potentially diverse lighting conditions, poses, ages, and camera angles.

Once reference images are collected, we use FaceNet \cite{Schroff_2015_facenet} to compute facial recognition features for each image. The features for each set of 10 reference images are averaged to create a general representation for each name. Figure \ref{pipeline} shows how FaceNet features are used in our model. At inference time, FaceNet features are also computed for a video's keyframes. We then take the cosine similarity between the representations for names appearing in the caption and the features for each keyframe in the video. In practice, these keyframe features are pre-computed for efficiency. The similarity scores are passed on to our model's classification head to be used alongside features from other modalities. 

This approach to person identification has a few drawbacks. The reference images of named entities from Google Images are not manually curated, which introduces issues such as the appearance of multiple people in a reference image. Additionally, in some cases, an individual might be referenced first by their full name, i.e. "Alice Appleseed," and then only by their first name, "Alice." Our NER approach does not account for this, and "Alice" would not be associated with "Alice Appleseed." In this case, the system may try to verify the appearance of "Alice" in a video, without knowing which "Alice" it should look for. 

This is less of a problem for individuals who are commonly referred to by a single name, or a variety of distinctive names. For instance, celebrities can often be uniquely identified by their first or last name, and many politicians are referred to by their last names. While there will be separate reference images for the named entities "Kanye West" and "Kanye," or the entities "Nancy Pelosi" and "Pelosi," they will be faithful representations of the same person.

\subsubsection{Name Verification}
While it is possible to verify the appearance of named entities from captions in videos, we can also compare captions to audio transcripts. 
This can alleviate the problem where an individual might be a topic of discussion, rather than a subject appearing in a video.

To accomplish this, we compute character-based embeddings for the names which appear in captions and/or transcripts. The intuition behind this operation is that our goal is to focus on misspellings, rather than any semantic concepts associated with names. 
Given a string representing a named entity, we convert each character in the string to its lower-case ASCII numerical value and pad to a maximum length of 64 characters. In our dataset, 100\% of strings identified as names have at most 64 characters. We then feed this vector into a 2-layer fully connected network, with hidden size 64 and output size 32.

These name embeddings are then passed on to our classification head for use along with other modalities, as shown in Figure \ref{pipeline}. By using learned embeddings, we are able to make comparisons between captions and audio transcripts, even when there are transcription errors in named entities.

\subsection{Facebook Reactions}
Since our data is collected from Facebook posts, we also have access to the Facebook reactions for each post. In Facebook, users are able to select the following reactions in response to a post: Like, Love, Wow, Haha, Sad, Angry, and Care. We hypothesize that these reactions can provide a coarse measure of the perceived semantics of an entire post, taking into consideration all of its modalities. In that case, the semantic inconsistency between an uplifting video paired with a sad or inflammatory caption might be reflected in an inconsistency between the post as a whole and its reactions.

We take the normalized reactions to a post as an input feature to our model, shown in Figure \ref{pipeline}. To normalize reactions, we divide the raw count of each reaction, such as Love, by the total number of reactions a post received. In this manner, viewers' reactions to content are separated from the popularity of a post, as all normalized reactions are bound between 0 and 1.
One problem with this approach is that our data is collected from 2010 to 2020, but reactions were first introduced in 2016, and the Care reaction was added in 2020. 
So, for some posts in our dataset, users would not have been able to choose a reaction other than Like. 

\subsection{Ensemble Feature Extraction}
We adopt a uni-modal ensemble approach to multi-modal fusion, as shown in Figure \ref{pipeline}.
To classify whether or not a post has a semantic inconsistency, we take as input the video split into clips starting at keyframes, the audio transcript, the normalized reactions to the video's pristine post, and a potentially inconsistent caption. In addition to the named entity verification features described in Section \ref{ssection:facialverification}, we compute features for the caption, transcript, and video clip inputs.

Both the audio transcript and caption are processed using a pre-trained BERT \cite{devlin2019bert} language model, implemented by HuggingFace \cite{wolf-etal-2020-transformers_huggingface}. 
When using the language model, inputs are truncated to their first 1024 characters, and split into two sets of characters with length 512. We need to create these splits because the maximum input length to the language model is 512 characters. In our dataset, 60\% of audio transcripts and 99.97\% of captions have at most 1024 characters.

The video clips are processed using both a video-understanding network and an object detection network. 
For video understanding, we use S3D-MIL-NCE (S3D) \cite{miech19endtoend}, and for object detection, we use a ResNet50 model \cite{he2015resnet}. S3D is run on the full 32-frame sequence in each of the video clips, while ResNet is run on the first frame of each clip. 
We use the \texttt{mixed\_5c} output of S3D.

\subsection{Multi-Modal Fusion}
For each modality, we learn an embedding to a shared semantic latent space. Figure \ref{pipeline} shows our full model architecture. 
Each embedding function is implemented as a 2-layer fully connected network, mapping from the output feature space of a modality's feature extraction network to a common 256-dimensional latent space. 
The learned semantic embeddings for video clips, object detection, audio transcript, and caption are concatenated and passed through a Long Short-Term Memory (LSTM) \cite{lstm} module to condense information from the clips into one summary feature vector. 
This fuses multi-modal content at the clip level, before the output of the LSTM is concatenated with named entity verification features. The final combined feature vector is passed on to our classification network.
Our classifier is implemented as a 3-layer fully connected network, with input size 1096, hidden layer sizes 512 and 128, and output size 2.

\section{Experiments}
\label{section:experiments}

\subsection{Experimental Design}
We train our model with the self-supervised dataset described in Section \ref{section:data}. We optimize the binary cross-entropy loss function, where our model classifies caption, audio transcript, and video appearance tuples as either pristine or inconsistent.

We report classification accuracy for our experiments, computed as the percentage of examples correctly identified as either pristine or inconsistent in our balanced test set. Our data is split such that 15\% of the examples are reserved for the test set, and the other 85\% for training and validation.

\subsection{Results and Ablation Studies}

\begin{table*}[h]
  \caption{Binary classification accuracy (\%) of heavily multi-modal models}
  \centering
  \begin{tabular}{lllllllll}
    \toprule
    \multicolumn{9}{c}{Modality or Feature Removed}                   \\
    \cmidrule(r){2-9}
    Model        & Names \& Faces & Caption & Names  & Video  & Transcript  & Faces  & Reacts & None \\
    \midrule
    Full         &$49.8$          & $54.2$  & $52.4$ & $54.7$ & $57.0$      & $56.9$ & $57.4$ & $58.3$    \\
    No OD        &$49.9$          & $51.5$  & $54.8$ & $56.5$ & $59.5$      & $59.6$ & $\mathbf{60.5}$ & $\mathbf{60.5}$    \\
    \bottomrule
  \end{tabular}
  \label{table:multimodal_experiments}
\end{table*}

\begin{table}[h]
  \caption{Best model confusion matrix (\%)}
  \centering
  \begin{tabular}{lll}
    \toprule
    & Predict Pristine & Predict Inconsistent \\
    
    \cmidrule(r){2-3}
    Pristine Examples  & $51.0$  & $49.0$  \\
    Inconsistent Examples  & $28.6$  & $71.4$  \\
    
    \bottomrule
  \end{tabular}
  \label{table:confusion-matrix}
\end{table}

\begin{table}[h]
  \caption{Binary classification accuracy (\%) of uni- and bi-modal models}
  \centering
  \begin{tabular}{llllllll}
    \toprule
    \multicolumn{5}{c}{Modalities Used}                   \\
    \cmidrule(r){1-5}
    Caption \& Video &  Video  & Caption & Faces   & Names \\
    \midrule
    $49.6$           & $49.8$  & $49.9$  & $51.7$  & $\mathbf{53.5}$      \\
    
    \bottomrule
  \end{tabular}
  \label{table:unimodal_experiments}
\end{table}

We perform a variety of ablation experiments to characterize the impact of each modality on the accuracy of our model. Results are shown below in Table \ref{table:multimodal_experiments}, with each modality removed one-by-one. Due to the fact that removing object detection features improved model performance, we perform one-by-one removal ablation studies again, with object detection features always removed. These experiments are referred to as "No OD" models in Table \ref{table:multimodal_experiments}. Note that "removing" a modality refers to removing its features or embeddings from our classifier. For instance, removing the video appearance makes the semantic video embeddings inaccessible to our classifier, although the video is still available for checking named entity consistency with facial verification. 

As seen in Table \ref{table:multimodal_experiments}, best performance is achieved by using all modalities, except object detection features, and reaches classification accuracy of 60.5\%. Table \ref{table:confusion-matrix} shows the confusion matrix for this model. We observe that the model is more accurate when classifying inconsistent examples. Specifically, it can correctly detect inconsistency 71\% of the time, and detects consistency 51\% of the time. Table \ref{table:unimodal_experiments} shows results for models using one or two modalities. 

We observe that named entity verification is key to model accuracy, as seen in Table \ref{table:multimodal_experiments}. Without facial verification, classification accuracy decreases slightly to 59.6\%. Without comparing names between captions and transcripts, classification accuracy falls to 54.8\%. Without performing either consistency check, classification accuracy falls to 49.9\%, essentially random.

We find that named entities are not the only useful information provided by captions. As seen in Table \ref{table:multimodal_experiments}, when semantic embeddings for captions are removed, accuracy falls to 54.2\% and 51.5\%, depending on whether or not object detection features are present, respectively. 
When caption embeddings are removed, the names present in the caption are still made available to our named entity verification process. Combination of semantic embeddings and named entity verification is the best use of information in the caption modality.

We note that video embeddings from S3D are more important than object detection (OD) embeddings from ResNet. In fact, removing OD embeddings improves the performance of our model, while removing S3D embeddings lowers performance. When OD embeddings are present, removing S3D embeddings leads to 3.8\% lower accuracy, and without OD embeddings, removing S3D embeddings leads to 4\% lower accuracy. For other datasets with instructional or cooking videos, we expect OD to play a more important role.

This could be due to the fact that features from S3D contain representations of objects, so the contribution of object detection features is diluted. OD features are not temporally aware, and so they cannot contain all the information represented in S3D features. Furthermore, the ResNet50 model we take features from is trained for image classification, which may be too general of a task to be useful for modelling abstract video semantics.

We note that Facebook reactions do not seem to provide a useful signal, as removing them from our model did not decrease performance.

Finally, we observe that multi-modal fusion is necessary for achieving the best possible accuracy. Removing any one of our modalities decreases performance, with the exception of reactions. More importantly, no uni-modal model can perform better than random. 
Accuracy for uni- and bi-modal models is shown in Table \ref{table:unimodal_experiments}. Caption-only and video-only models achieve 49.9\% and 49.8\% classification accuracy, respectively, confirming that our dataset does not have linguistic or visual bias. 
A model combining caption and video clip embeddings achieves 49.6\% accuracy, highlighting the importance of incorporating additional modalities and features.
A model which solely compares named entities in captions and audio transcripts achieves 53.5\% accuracy, and a model which compares named entities in captions with video frame facial verification features achieves 51.7\% accuracy. While attending to named entities is important, named entities alone are not sufficient for our model to achieve the highest possible accuracy.

\section{Conclusion}
\label{section:conclusion}
We have introduced a novel multi-modal semantic inconsistency detection system, along with a 4k-large dataset for self-supervising semantic alignment detection in real-world social media posts. 
We demonstrate the importance of making use of modalities beyond video appearance and captions, including transcription, facial verification, and possibly misspelled named entity comparison.

We observe that fusion across modalities is key to detecting semantic inconsistencies. 
We find that named entities provide strong signals for verifying consistency across modalities, and that verifying named entities using both language-based and visual methods is better than only using one. Semantic consistency checks cannot be fully explained by named entity verification, however, highlighting the need to consider semantic embeddings for language and video.

Future work could explore aspects of attributing and characterizing inconsistencies. 
Modules for explainable facial verification and author attribution could take steps towards addressing this. Our approach would likely benefit from more data, and we are interested in expanding data collection to other social networks such as Twitter and TikTok. 
Increasing the size of our dataset might also allow for more challenging inconsistencies during training time.

\newpage
\printbibliography

@inproceedings{miech19howto100m,
   title={How{T}o100{M}: {L}earning a {T}ext-{V}ideo {E}mbedding by {W}atching {H}undred {M}illion {N}arrated {V}ideo {C}lips},
   author={Miech, Antoine and Zhukov, Dimitri and Alayrac, Jean-Baptiste and Tapaswi, Makarand and Laptev, Ivan and Sivic, Josef},
   booktitle={ICCV},
   year={2019},
}

@article{nytrobotnews,
    author={Jaclyn Peiser},
    date={2019-02-05},
    title={The Rise of the Robot Reporter},
    journal={The New York Times},
    url={https://www.nytimes.com/2019/02/05/business/media/artificial-intelligence-journalism-robots.html},
    
}

@inproceedings{hara3dcnns,
  author={Kensho Hara and Hirokatsu Kataoka and Yutaka Satoh},
  title={Can Spatiotemporal 3D CNNs Retrace the History of 2D CNNs and ImageNet?},
  booktitle={Proceedings of the IEEE Conference on Computer Vision and Pattern Recognition (CVPR)},
  pages={6546--6555},
  year={2018},
}

@INPROCEEDINGS{inception3d,
  author={Zhao, Chen and Han, Jungang and Jia, Yang},
  booktitle={2017 International Conference on Computational Science and Computational Intelligence (CSCI)}, 
  title={3D Inception Convolutional Neural Networks for Automatic Lung Nodule Detection}, 
  year={2017},
  volume={},
  number={},
  pages={1649-1653},
  doi={10.1109/CSCI.2017.287}}

@inproceedings{miech19endtoend,
   title={{E}nd-to-{E}nd {L}earning of {V}isual {R}epresentations from {U}ncurated {I}nstructional {V}ideos},
   author={Miech, Antoine and Alayrac, Jean-Baptiste and Smaira, Lucas and Laptev, Ivan and Sivic, Josef and Zisserman, Andrew},
   booktitle={CVPR},
   year={2020},
}

@misc{brown2020language_gpt3,
      title={Language Models are Few-Shot Learners}, 
      author={Tom B. Brown and Benjamin Mann and Nick Ryder and Melanie Subbiah and Jared Kaplan and Prafulla Dhariwal and Arvind Neelakantan and Pranav Shyam and Girish Sastry and Amanda Askell and Sandhini Agarwal and Ariel Herbert-Voss and Gretchen Krueger and Tom Henighan and Rewon Child and Aditya Ramesh and Daniel M. Ziegler and Jeffrey Wu and Clemens Winter and Christopher Hesse and Mark Chen and Eric Sigler and Mateusz Litwin and Scott Gray and Benjamin Chess and Jack Clark and Christopher Berner and Sam McCandlish and Alec Radford and Ilya Sutskever and Dario Amodei},
      year={2020},
      eprint={2005.14165},
      archivePrefix={arXiv},
      primaryClass={cs.CL}
}

@article{radford2019language_gpt2,
  title={Language Models are Unsupervised Multitask Learners},
  author={Radford, Alec and Wu, Jeff and Child, Rewon and Luan, David and Amodei, Dario and Sutskever, Ilya},
  year={2019}
}

@misc{vaswani2017attention,
      title={Attention Is All You Need}, 
      author={Ashish Vaswani and Noam Shazeer and Niki Parmar and Jakob Uszkoreit and Llion Jones and Aidan N. Gomez and Lukasz Kaiser and Illia Polosukhin},
      year={2017},
      eprint={1706.03762},
      archivePrefix={arXiv},
      primaryClass={cs.CL}
}

@article{lstm, 
author = {Hochreiter, Sepp and Schmidhuber, J\"{u}rgen}, title = {Long Short-Term Memory}, year = {1997}, issue_date = {November 15, 1997}, publisher = {MIT Press}, address = {Cambridge, MA, USA}, volume = {9}, number = {8}, issn = {0899-7667}, url = {https://doi.org/10.1162/neco.1997.9.8.1735}, doi = {10.1162/neco.1997.9.8.1735}, journal = {Neural Comput.}, month = nov, pages = {1735–1780}, numpages = {46} }

@misc{he2015resnet,
      title={Deep Residual Learning for Image Recognition}, 
      author={Kaiming He and Xiangyu Zhang and Shaoqing Ren and Jian Sun},
      year={2015},
      eprint={1512.03385},
      archivePrefix={arXiv},
      primaryClass={cs.CV}
}

@article{Schroff_2015_facenet,
   title={FaceNet: A unified embedding for face recognition and clustering},
   ISBN={9781467369640},
   url={http://dx.doi.org/10.1109/CVPR.2015.7298682},
   DOI={10.1109/cvpr.2015.7298682},
   journal={2015 IEEE Conference on Computer Vision and Pattern Recognition (CVPR)},
   publisher={IEEE},
   author={Schroff, Florian and Kalenichenko, Dmitry and Philbin, James},
   year={2015},
   month={Jun}
}

@misc{hannun2014deepspeech,
      title={Deep Speech: Scaling up end-to-end speech recognition}, 
      author={Awni Hannun and Carl Case and Jared Casper and Bryan Catanzaro and Greg Diamos and Erich Elsen and Ryan Prenger and Sanjeev Satheesh and Shubho Sengupta and Adam Coates and Andrew Y. Ng},
      year={2014},
      eprint={1412.5567},
      archivePrefix={arXiv},
      primaryClass={cs.CL}
}

@software{youtubedl,
  author = {youtube-dl Developers},
  title = {{youtube-dl}},
  year = 2021,
  version = {2021.01.24.1},
  url = {https://youtube-dl.org}
}

@software{spacy,
  author = {Honnibal, Matthew and Montani, Ines and Van Landeghem, Sofie and Boyd, Adriane},
  title = {{spaCy: Industrial-strength Natural Language Processing in Python}},
  year = 2020,
  publisher = {Zenodo},
  doi = {10.5281/zenodo.1212303},
  url = {https://doi.org/10.5281/zenodo.1212303}
}

@software{ffmpeg,
    author = {{FFmpeg Developers}},
    version = {4.3.1},
    date = {2020},
    url = {http://ffmpeg.org/}
}

@misc{liu2019roberta,
      title={RoBERTa: A Robustly Optimized BERT Pretraining Approach}, 
      author={Yinhan Liu and Myle Ott and Naman Goyal and Jingfei Du and Mandar Joshi and Danqi Chen and Omer Levy and Mike Lewis and Luke Zettlemoyer and Veselin Stoyanov},
      year={2019},
      eprint={1907.11692},
      archivePrefix={arXiv},
      primaryClass={cs.CL}
}

@misc{devlin2019bert,
      title={BERT: Pre-training of Deep Bidirectional Transformers for Language Understanding}, 
      author={Jacob Devlin and Ming-Wei Chang and Kenton Lee and Kristina Toutanova},
      year={2019},
      eprint={1810.04805},
      archivePrefix={arXiv},
      primaryClass={cs.CL}
}

@misc{su2020vlbert,
      title={VL-BERT: Pre-training of Generic Visual-Linguistic Representations}, 
      author={Weijie Su and Xizhou Zhu and Yue Cao and Bin Li and Lewei Lu and Furu Wei and Jifeng Dai},
      year={2020},
      eprint={1908.08530},
      archivePrefix={arXiv},
      primaryClass={cs.CV}
}

@misc{lu2019vilbert,
      title={ViLBERT: Pretraining Task-Agnostic Visiolinguistic Representations for Vision-and-Language Tasks}, 
      author={Jiasen Lu and Dhruv Batra and Devi Parikh and Stefan Lee},
      year={2019},
      eprint={1908.02265},
      archivePrefix={arXiv},
      primaryClass={cs.CV}
}

@misc{li2019unicodervl,
      title={Unicoder-VL: A Universal Encoder for Vision and Language by Cross-modal Pre-training}, 
      author={Gen Li and Nan Duan and Yuejian Fang and Ming Gong and Daxin Jiang and Ming Zhou},
      year={2019},
      eprint={1908.06066},
      archivePrefix={arXiv},
      primaryClass={cs.CV}
}

@misc{bertasius2021spacetime,
      title={Is Space-Time Attention All You Need for Video Understanding?}, 
      author={Gedas Bertasius and Heng Wang and Lorenzo Torresani},
      year={2021},
      eprint={2102.05095},
      archivePrefix={arXiv},
      primaryClass={cs.CV}
}

@misc{xie2018rethinking,
      title={Rethinking Spatiotemporal Feature Learning: Speed-Accuracy Trade-offs in Video Classification}, 
      author={Saining Xie and Chen Sun and Jonathan Huang and Zhuowen Tu and Kevin Murphy},
      year={2018},
      eprint={1712.04851},
      archivePrefix={arXiv},
      primaryClass={cs.CV}
}

@inproceedings{tan2019lxmert,
  title={LXMERT: Learning Cross-Modality Encoder Representations from Transformers},
  author={Tan, Hao and Bansal, Mohit},
  booktitle={Proceedings of the 2019 Conference on Empirical Methods in Natural Language Processing},
  year={2019}
}

@misc{crowdtangle,
      title={CrowdTangle}, 
      author={{CrowdTangle Team}},
      year={2021},
      address={Facebook, Menlo Park, California, United States}
}

@MISC{wikipedia:bridgers,
    author = {David Lee},
    title = {Phoebe Bridgers (41599189180) (cropped).jpg},
    year = {2018},
    note = {[Online; accessed May 7, 2021]},
    url = {https://commons.wikimedia.org/wiki/File:Phoebe_Bridgers_(41599189180)_(cropped).jpg}
}

@MISC{wikipedia:mccartney,
    author = {{Raph\_PH}},
    title = {Paul McCartney in October 2018.jpg},
    year = {2018},
    note = {[Online; accessed May 7, 2021]},
    url = {https://commons.wikimedia.org/wiki/File:Paul_McCartney_in_October_2018.jpg}
}

@inproceedings{wolf-etal-2020-transformers_huggingface,
    title = "Transformers: State-of-the-Art Natural Language Processing",
    author = "Thomas Wolf and Lysandre Debut and Victor Sanh and Julien Chaumond and Clement Delangue and Anthony Moi and Pierric Cistac and Tim Rault and Rémi Louf and Morgan Funtowicz and Joe Davison and Sam Shleifer and Patrick von Platen and Clara Ma and Yacine Jernite and Julien Plu and Canwen Xu and Teven Le Scao and Sylvain Gugger and Mariama Drame and Quentin Lhoest and Alexander M. Rush",
    booktitle = "Proceedings of the 2020 Conference on Empirical Methods in Natural Language Processing: System Demonstrations",
    month = oct,
    year = "2020",
    address = "Online",
    publisher = "Association for Computational Linguistics",
    url = "https://www.aclweb.org/anthology/2020.emnlp-demos.6",
    pages = "38--45"
}

@inproceedings{wang2019cnngenerated,
  title={CNN-generated images are surprisingly easy to spot...for now},
  author={Wang, Sheng-Yu and Wang, Oliver and Zhang, Richard and Owens, Andrew and Efros, Alexei A},
  booktitle={CVPR},
  year={2020}
}

@misc{luo2021newsclippings,
      title={NewsCLIPpings: Automatic Generation of Out-of-Context Multimodal Media}, 
      author={Grace Luo and Trevor Darrell and Anna Rohrbach},
      year={2021},
      eprint={2104.05893},
      archivePrefix={arXiv},
      primaryClass={cs.CV}
}

@inproceedings{zellers2019grover,
    title={Defending Against Neural Fake News},
    author={Zellers, Rowan and Holtzman, Ari and Rashkin, Hannah and Bisk, Yonatan and Farhadi, Ali and Roesner, Franziska and Choi, Yejin},
    booktitle={Advances in Neural Information Processing Systems 32},
    year={2019}
}

@inproceedings{shekhar-etal-2017-foil,
    title = "{FOIL} it! Find One mismatch between Image and Language caption",
    author = "Shekhar, Ravi  and
      Pezzelle, Sandro  and
      Klimovich, Yauhen  and
      Herbelot, Aur{\'e}lie  and
      Nabi, Moin  and
      Sangineto, Enver  and
      Bernardi, Raffaella",
    booktitle = "Proceedings of the 55th Annual Meeting of the Association for Computational Linguistics (Volume 1: Long Papers)",
    month = jul,
    year = "2017",
    address = "Vancouver, Canada",
    publisher = "Association for Computational Linguistics",
    url = "https://www.aclweb.org/anthology/P17-1024",
    doi = "10.18653/v1/P17-1024",
    pages = "255--265",
}

@Article{Luo2020UniVL,
  author  = {Huaishao Luo and Lei Ji and Botian Shi and Haoyang Huang and Nan Duan and Tianrui Li and Jason Li and Taroon Bharti and Ming Zhou},
  title   = {UniVL: A Unified Video and Language Pre-Training Model for Multimodal Understanding and Generation},
  journal = {arXiv preprint arXiv:2002.06353},
  year    = {2020},
}

@ARTICLE{habibian_video2vec,
  author={Habibian, Amirhossein and Mensink, Thomas and Snoek, Cees G. M.},
  journal={IEEE Transactions on Pattern Analysis and Machine Intelligence}, 
  title={Video2vec Embeddings Recognize Events When Examples Are Scarce}, 
  year={2017},
  volume={39},
  number={10},
  pages={2089-2103},
  doi={10.1109/TPAMI.2016.2627563}}

@InProceedings{tanDIDAN2020,
     author={Reuben Tan and Bryan A. Plummer and Kate Saenko},
     title={Detecting Cross-Modal Inconsistency to Defend Against Neural Fake News},
     booktitle={Empirical Methods in Natural Language Processing (EMNLP)},
     year={2020} }

@inproceedings{li-etal-2020-hero,
    title = "{HERO}: Hierarchical Encoder for {V}ideo+{L}anguage Omni-representation Pre-training",
    author = "Li, Linjie  and
      Chen, Yen-Chun  and
      Cheng, Yu  and
      Gan, Zhe  and
      Yu, Licheng  and
      Liu, Jingjing",
    booktitle = "Proceedings of the 2020 Conference on Empirical Methods in Natural Language Processing (EMNLP)",
    month = nov,
    year = "2020",
    address = "Online",
    publisher = "Association for Computational Linguistics",
    url = "https://www.aclweb.org/anthology/2020.emnlp-main.161",
    doi = "10.18653/v1/2020.emnlp-main.161",
    pages = "2046--2065",
}

\end{document}